\documentclass[twoside,a4paper]{llncs}
\usepackage{graphicx}
\usepackage{subcaption}
\usepackage{caption}
\usepackage{hhline}
\usepackage{siunitx}
\usepackage{amsmath}
\usepackage{amssymb}
\usepackage{multirow}
\usepackage{blindtext, xcolor}

\begin{document}
\mainmatter
\title{Active Learning for Segmentation by Optimizing Content Information for Maximal Entropy}
%
\author{Firat Ozdemir$^*$ \and Zixuan Peng$^*$ \and Christine Tanner$^*$  \and Philipp Fuernstahl$^\dagger$ \and Orcun Goksel$^*$}
\institute{$^*$Computer-assisted Applications in Medicine, ETH Zurich, Switzerland\\
$^\dagger$CARD, 
University Hospital Balgrist, University of Zurich, Switzerland} 

\maketitle

\begin{abstract}

Segmentation is essential for medical image analysis tasks such as intervention planning, therapy guidance, diagnosis, treatment decisions. Deep learning is becoming increasingly prominent for segmentation, where the lack of annotations, however, often becomes the main limitation.
Due to privacy concerns and ethical considerations, most medical datasets are created, curated, and allow access only locally.
Furthermore, current deep learning methods are often suboptimal in translating anatomical knowledge between different medical imaging modalities.
Active learning can be used to select an informed set of image samples to request for manual annotation, in order to best utilize the limited annotation time of clinical experts for optimal outcomes, which we focus on in this work. 
Our contributions herein are two fold: (1) we enforce domain-representativeness of selected samples using a proposed penalization scheme to maximize information at the network \textit{abstraction layer}, and (2) we propose a Borda-count based sample querying scheme for selecting samples for segmentation.
Comparative experiments with baseline approaches show that the samples queried with our proposed method, where both above contributions are combined, result in significantly improved segmentation performance for this active learning task.
\end{abstract}

\section{Introduction}
\label{sec:introduction}

Segmentation has several medical applications, such as patient-specific surgical planning. 
Due to limited resources of expert physicians, detailed manual annotations are often not possible, even when desired anatomy may be visible with sufficient contrast using non-invasive imaging modalities such as MRI and ultrasound. 
Deep learning has shown encouraging performance for segmentation~\cite{baumgartner2017exploration,Chen2016DCAN}, but often only when sufficient amount of labeled data for a target anatomy is available. 
Medical image data across different medical centers is often not uniform, for instance with respect to machine manufacturer, imaging settings, and cohort demographics. Thus, studies and corresponding annotations are only carried out in isolated datasets, with difficulties in merging information with data sharing, patient rights, and confidentiality concerns.
Hence, a sufficiently large dataset for a given task needs to be labeled.
\textit{Active learning} aims at maximizing the prediction performance through an intelligent sample querying system so that the limited expert annotation resources can be properly managed as opposed to training on a randomly selected next batch of samples which would contain a lot of redundancy.
In a clinical environment, one can imagine that expert(s) will allocate a fixed amount of annotation time per time interval (i.e.,\ week), hence the correct use of this time (i.e.,\ on most valuable samples) is essential. 
Therefore, the segmentation framework would be initially provided a very limited labeled dataset, which will be extended with a certain batch size of samples intelligently selected at each iteration of the active learning.

Intuitively, the prediction confidence of a learned model can be used as a surrogate metric for its potential accuracy, in order to propose the most \emph{uncertain} predictions for future manual annotation.
In~\cite{gal2015dropout}, \textit{MC dropout} is proposed to sample from the approximate trained model posterior, which can be used to quantify an \textit{uncertainty} metric through variations in the model predictions for a given input. 
Based on this, several approaches of querying the next batch of data are studied and compared with uniform random sampling in~\cite{gal2017deep}.
Unfortunately, it is intractable to assess conditional uncertainty of multiple samples; e.g.\, would $i^\mathrm{th}$ sample be still as uncertain as before once $j^\mathrm{th}$ sample is queried and trained for.
Thus, it is intuitive to select a \textit{representative} subset of these uncertain samples to reduce redundancy. 
Using a simplified version of DCAN~\cite{Chen2016DCAN} architecture (which has won the first place in the 2015 MICCAI Gland Segmentation Challenge~\cite{Sirinukunwattana2017gland}) for the purpose of faster training, a state-of-the-art method was proposed in~\cite{yang2017suggestive} to select optimal sample images to annotate. 
First, a batch of \emph{uncertain} samples is chosen based on the mean variance of multiple network predictions, followed by picking a subset of these using \emph{maximum set coverage}~\cite{Feige1998a} over the \textit{image descriptors} of these samples.
Recently in~\cite{ozdemir2018learn}, a \textit{content distance}~\cite{gatys2016image} concept was proposed to quantify the similarity between two images, for selecting representative samples in class-incremental learning.

Herein we propose two main novelties for querying samples at an active learning step:
(1)~we add an additional constraint on the \textit{abstraction layer}~\cite{ozdemir2018learn} activations during training to maximize information content at this level.
We show that this additional constraint improves sample suitability that boosts segmentation performance from active learning.
(2)~Instead of the two step sample querying procedure (i.e., first select based on \textit{uncertainty}, then cull using \textit{representativeness}), we propose a Borda-count based method.
This alone provides improvement over the state-of-the-art~\cite{yang2017suggestive}; and when used in conjunction with our novel constraint above, it yields even further segmentation improvement.

\section{Estimating Surrogate Metrics for Representativeness}
\label{sec:representativeness_metrics}

\textbf{Background.} 
In~\cite{yang2017suggestive}, multiple FCNs were trained to estimate uncertainty for a given image through variation in their inferences. 
To make the FCN predictions diverse, the annotated dataset was also bootstrapped when training each model.
However, training several models is a costly operation and with larger number of models, one should bootstrap a smaller portion of the already-minimal dataset available in the early stages of typical active learning scenarios.

\begin{figure}[t]
    \centering
    \includegraphics[width = 0.99\textwidth]{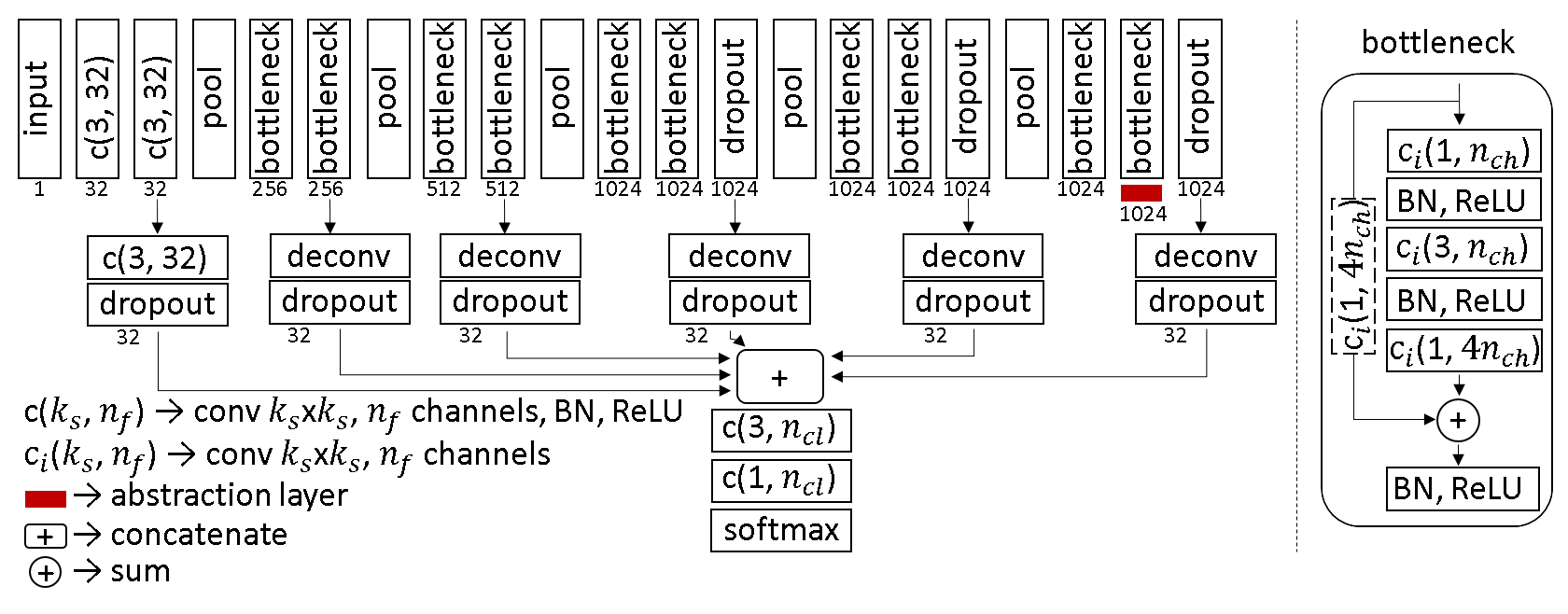}
    \caption{DCAN network for \textit{Suggestive\,Annotation} with additional spatial dropout layers. $n_{ch}$ is the number of filters in respective block, BN is batch normalization, and $n_{cl}$ is the number of classes. In consecutive bottlenecks, the first uses convolution filter in shortcuts to match tensor size while the second does not.}
    \label{fig:mod_DCAN_architecture}
\end{figure}

\emph{In our work,} as a baseline, we implemented an improved version of the \textit{Suggestive Annotation} framework~\cite{yang2017suggestive}. 
We added dropout layers (c.f.\ Fig.\ref{fig:mod_DCAN_architecture}) to allow for MC dropout~\cite{gal2015dropout}, through which one can compute the voxel-wise variance across $n_i$ inferences, and average it over all input voxels.
The first step in querying samples is to pick the most uncertain $n_\mathrm{unc}$ samples $S_\mathrm{unc}$ from the set of non-annotated data $D_\mathrm{pool}$.
For representativeness, ``image descriptor'' $I_i^c$ of every image $I_i \in D_\mathrm{pool}$ is computed as described in~\cite{yang2017suggestive} at the abstraction layer, $l_\mathrm{abst}$ (c.f. Fig.~\ref{fig:mod_DCAN_architecture}).
Using cosine similarity $d_\mathrm{sim}(I_i, I_j) = \cos(I_i^c, I_j^c)$ between the descriptors of images $I_i$ and $I_j$, the maximum set-cover~\cite{Feige1998a} over $D_\mathrm{pool}$ is computed using descriptors from $S_\mathrm{unc}$ for the top $n_\mathrm{rep}$ images.
We call this method of using uncertainty and the above image descriptor (ID) as UNC-ID hereafter.

\noindent\textbf{Content Distance.} 
The image descriptor $I_i^c$ averages the spatial information at the corresponding layer activations. 
While this allows for a spatially invariant means of representing a given image at a very abstract level, higher order features extracted at this stage would be blurred by this process.
Alternatively, layer activation responses $R^l(I_i)$ of a pretrained classification network at a layer $l$ can be used to describe the content of an image $I_i$~\cite{gatys2016image}.
Then, content distance ($d_\mathrm{cont}$) between images $I_i$ and $I_j$ is defined as the mean squared error between their responses at layer $l$:
\begin{equation}
	d_\mathrm{cont}(I_i,I_j) = \frac{1}{N}\sum^N (R^{l}(I_i) - R^{l}(I_j))^2
    \label{eq:dcont}
\end{equation}
A similar notion can be applied to active learning problems, where input images are described by the activation response at the $l_\mathrm{abst}$ of the currently trained network (c.f. Fig.~\ref{fig:mod_DCAN_architecture}).

\noindent \textbf{Encoding Representativeness by Maximizing Entropy.}
Content distance defined in Eq.~(\ref{eq:dcont}) allows for finer content discrimination than image descriptors~\cite{yang2017suggestive}.
However, it has been suggested that activations at a single layer may not be sufficient for accurate content description~\cite{ozdemir2018learn}.
This is likely to particularly apply to segmentation networks, since network weights until $l_\mathrm{abst}$ are not optimized to describe the input image.
Therefore, it has been proposed to stack activations from multiple layers.
For a typical segmentation network, storing all layer activations of $D_\mathrm{pool}$ can quickly diverge to an unfeasible size. 
Alternatively, one can try to increase information content at the $l_\mathrm{abst}$ through maximizing its activation entropy~\cite{shannon2001a} along the feature channels.
Entropy loss can then be defined as: 
\begin{equation}
L_\mathrm{ent} = -\sum_x \mathrm{H}(R^{(l_\mathrm{abst}, x)})
\label{eq:entropy_loss}
\end{equation}
where $R^{(l_\mathrm{abst}, x)}$ are the input activations of all channels for spatial location $x$, and $x$ iterates over the width and height of the layer $l_\mathrm{abst}$.
Hence, total loss for the trained network becomes $L_\mathrm{total} = L_\mathrm{seg} + \lambda  L_\mathrm{ent}$, where $L_\mathrm{seg}$ is the segmentation loss, and $\lambda$ is used to scale the entropy loss $L_\mathrm{ent}$. 

Optimization of the network weights through entropy maximization is a novel regularization.
$L_\mathrm{ent}$ alone would have a tendency to alter network weights to only increase information, which may also encourage randomness.
With an appropriate $\lambda$, the network is forced to optimize parameters for the segmentation task while also increasing ``useful'' information content at the abstraction layer; as opposed to producing just noise at $l_\mathrm{abst}$.
Hence, additional content description for a given image can be retrieved from a single layer activation, making it a feasible alternative.
We refer to this method, where an entropy-based content distance (\textbf{E}CD) is used, as UNC-\textbf{E}CD.

\section{Sample Selection Strategy}
\label{sec:sample_selection}

For active learning, one should emphasize that the initial data size can be very small.
Until the model parameters are optimized for a sufficient coverage of the data distribution, the defined ``uncertainty'' metric might be misleading.
As a result, one can explore different ways to combine multiple metrics when querying samples instead of the conventional 2-step process.
An intuitive way to combine two metrics $m_k$ and $m_l$ would be to use $w_k m_k + w_l m_l$, where $w_k, w_l$ are weights.
However, \textit{uncertainty} and \textit{representativeness} metrics defined in Sec.~\ref{sec:representativeness_metrics} are not linearly combinable, even if normalized, due to non-linear unit increments. 
Therefore, we propose to use Borda count, where samples are ranked for each metric, and the next query sample $I_{i^*}$ is picked based on the best combined rank:
\begin{equation}
i^* = \arg\min_i(\sum_{m_k \in S_m} f_\mathrm{rank}(m_k(I_i)))
\label{eq:rank}
\end{equation}
where $S_m$ is the set of metrics $m_k$ to combine, and the $f_\mathrm{rank}$ function sorts the images based on the metric $m_k$.
When we use the ranking in Eq.~(\ref{eq:rank}) for samples selection, we denote this in our results with ``+'', e.g.\ content distance with uncertainty is named UNC+\textbf{E}CD.
In an active learning framework, the methods mentioned until now can be denoted as UNC+ID, UNC+\textbf{E}CD for ranking based sample selection and UNC-ID, UNC-\textbf{E}CD for uncertainty selection followed by representativeness selection.

\section{Experiments and  Results}
\label{sec:Experiments}

\begin{table}[t]
\centering
\caption{Dataset configuration}
\label{tbl:dataset}
\begin{tabular}{r|r|r|r|r|r|r|r}
\multicolumn{1}{c|}{Config} & \multicolumn{1}{c|}{\#volumes} & \multicolumn{1}{c|}{Left/Right} & \multicolumn{1}{c|}{vox res.\,{[}mm{]}} & \multicolumn{1}{c|}{image size\,{[}px{]}} & \multicolumn{1}{c|}{TR\,{[}s{]}} & \multicolumn{1}{c|}{TE\,{[}s{]}} & \multicolumn{1}{c}{FA\,[\si{\degree}]} \\ \hline
1                           & 20  & 9/11                           & 0.91 $\times$ 0.91 $\times$ 3.0                      & 192 $\times$ 192 $\times$ 64                             & 20                              & 1.70                            & 10                                                                        \\ \hline
2                           & 16 & 8/8                            & 0.83 $\times$ 0.83 $\times$ 3.0                      & 144 $\times$ 144 $\times$ 56                             & 20                              & 2.39                            & 10                                                                       
\end{tabular}
\end{table}

\begin{figure}[]
\centering
	\begin{subfigure}[b]{0.499\textwidth}
    \centering
        \includegraphics[width=0.999\textwidth, trim= 0cm 0.0cm 1.1cm 0cm, clip]{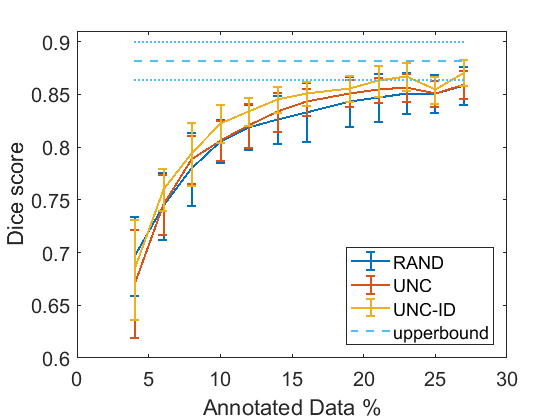}
        \caption{Dice score}
        \label{fig:scores:dice}
    \end{subfigure}%
    \begin{subfigure}[b]{0.499\textwidth}
    \centering
        \includegraphics[width=0.999\textwidth, trim= 0cm 0.0cm 1.1cm 0cm, clip]{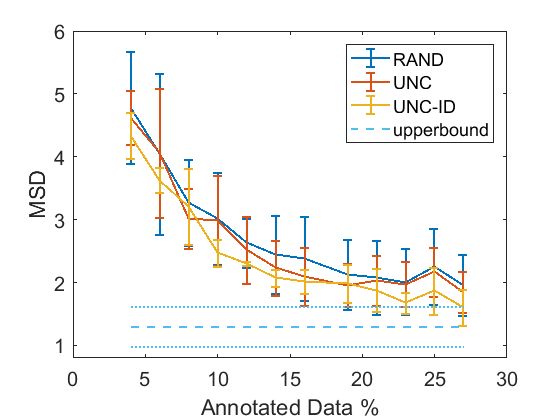}
        \caption{MSD [mm]}%
        \label{fig:scores:assd}
    \end{subfigure}%
   	\caption{Comparison between our implementation of the baseline method (UNC-ID) with random sampling (RAND) and only uncertainty-based (UNC) active learning methods.
	Training on 100\% of the data ($D_\mathrm{pool}$) is shown as upperbound.
    (a) Mean Dice score and (b) mean surface distance (MSD) with error bars covering the standard deviation of 5 hold-out experiments at every evaluation point.
    }
    \label{fig:scores}
\end{figure}

\begin{figure}[]
\centering
	\begin{subfigure}[b]{0.499\textwidth}
    \centering
        \includegraphics[width=0.999\textwidth, trim= 0cm 0.0cm 1.1cm 0cm, clip]{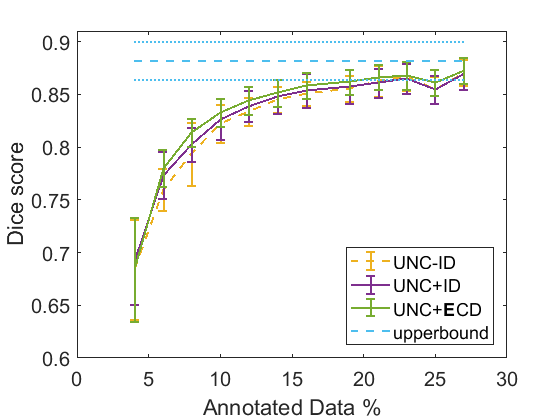}
        \caption{Dice score}
        \label{fig:scores2:dice}
    \end{subfigure}%
    \begin{subfigure}[b]{0.499\textwidth}
    \centering
        \includegraphics[width=0.999\textwidth, trim= 0cm 0.0cm 1.1cm 0cm, clip]{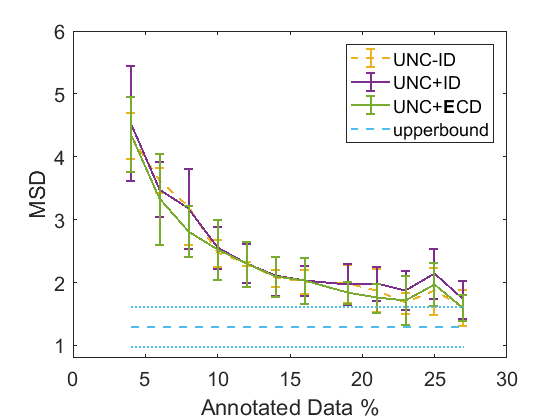}
        \caption{MSD [mm]}
        \label{fig:scores2:assd}
    \end{subfigure}%
   	\caption{Comparison of the baseline method (UNC-ID) with ranking based sample selection (UNC+ID) and the combination of our proposed extensions (UNC+\textbf{E}CD).
    Training on 100\% of the data ($D_\mathrm{pool}$) is shown as upperbound.
    (a) Mean Dice score and (b) mean surface distance (MSD) with error bars covering the standard deviation of 5 hold-out experiments at every evaluation point.
    The mean Dice score of UNC+\textbf{E}CD was statistically significantly higher than the baseline in 4 of 5 experiments (one-sided paired t-test at the 0.05 level).}
    \label{fig:scores2}
\end{figure}

\begin{figure}[t]
\centering
	\begin{subfigure}[b]{0.35\textwidth}
    \centering
        \includegraphics[width=0.98\textwidth, trim= 0cm 0.0cm 0cm 0cm, clip]{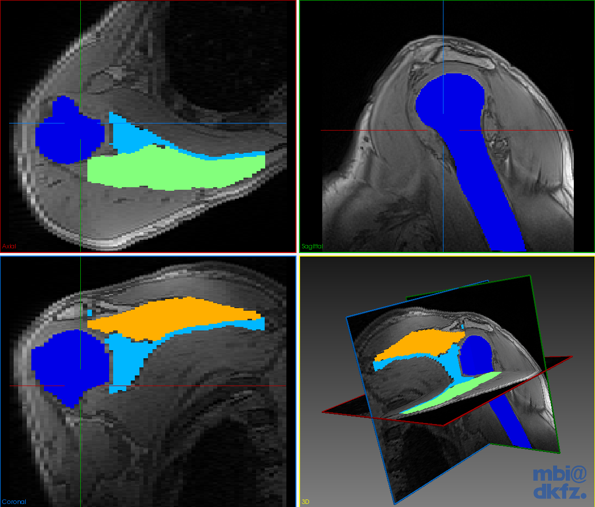}
        \caption{Gold Standard (GS)}
        \label{fig:qualRes:GS}
    \end{subfigure}%
    	\begin{subfigure}[b]{0.324\textwidth}
    \centering
    	\includegraphics[width=0.98\textwidth, trim= 0cm 0.0cm 0cm 0cm, clip]{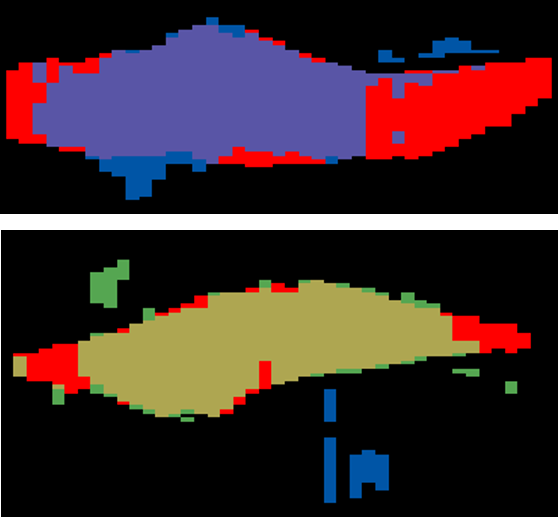}
        \caption{GS+Baseline}
        \label{fig:qualRes:overlay_baseline}
    \end{subfigure}%
    	\begin{subfigure}[b]{0.324\textwidth}
    \centering
    \includegraphics[width=0.98\textwidth, trim= 0cm 0.0cm 0cm 0cm, clip]{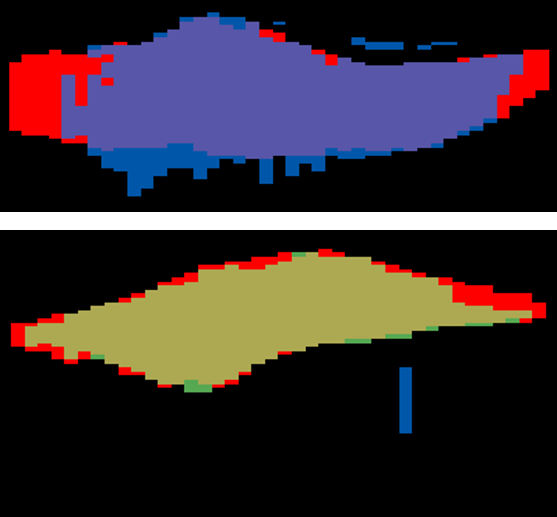}
        \caption{GS+Proposed}
        \label{fig:qualRes:overlay_proposed}
    \end{subfigure}\\%
    \begin{subfigure}[b]{0.495\textwidth}
    \centering
        \includegraphics[width=0.98\textwidth, trim= 0cm 0.0cm 0cm 0cm, clip]{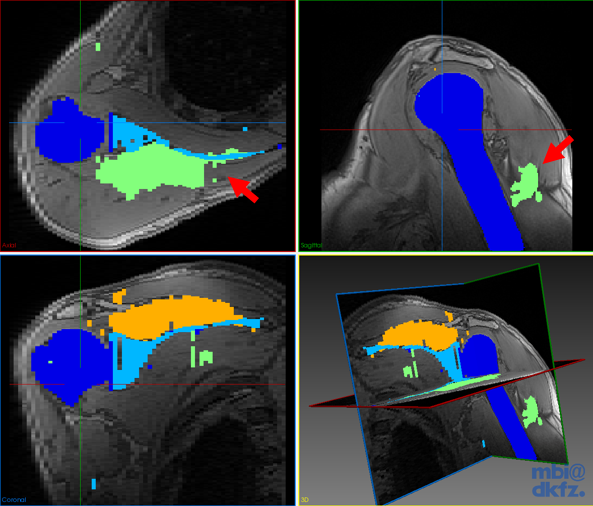}
        \caption{Baseline}
        \label{fig:qualRes:baseline}
    \end{subfigure}\hfill
    \begin{subfigure}[b]{0.495\textwidth}
    \centering
        \includegraphics[width=0.98\textwidth, trim= 0cm 0.0cm 0cm 0cm, clip]{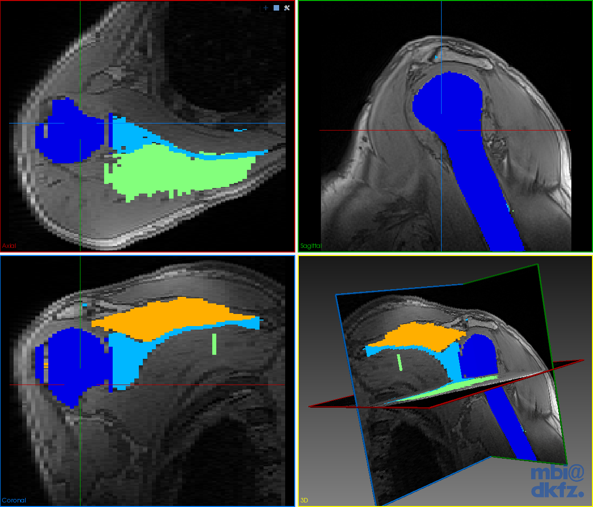}
        \caption{Proposed}
        \label{fig:qualRes:proposed}
    \end{subfigure}%
\caption{Segmentation of a test volume comparing baseline (UNC-ID) with proposed method (UNC+\textbf{E}CD)  after the first active learning step.
Segmentation of two muscles overlaid on GS annotation (red) for (b) baseline and (c) proposed method.
(d) Some of the substantial differences are pointed out by red arrows.
}
    \label{fig:qualRes}
\end{figure}

We have conducted experiments on an MR dataset of 36 patients diagnosed with rotator cuff tear (shoulders) according to specifications shown on Table~\ref{tbl:dataset}.
In an effort to regularize the dataset, Config2 images have been resized to match the voxel resolution of Config1, and then zero padded to match the in-plane image size of Config1.
The data has expert annotations of two bones (humerus \& scapula) and two muscle groups (supraspinatus \& infraspinatus\,+\,teres minor). 
Experiments have been conducted using NVIDIA Titan X GPU and Tensorflow library~\cite{tensorflow2015whitepaper}.

For all compared methods, we have used the modified DCAN architecture shown in Fig.\ref{fig:mod_DCAN_architecture}, training it on 2D in-plane slices with the parameters $n_{ch}$\,$=$\,$32$ and Adam optimizer.
When training the networks, learning rate of $5\times 10^{-4}$, dropout rate of 0.5, $n_i$$=$$17$, and minibatch size of 8 images were applied. 
At each active learning stage, including the initial training, models were trained for 8000 steps, which took about 48\,mins. 
Uncertainty metric is aggregated over the foreground classes to represent their mean uncertainty.
We used cross-entropy loss at the softmax layer (c.f. Fig.~\ref{fig:mod_DCAN_architecture}) for the $L_\mathrm{seg}$. 
Weight $\lambda$ for scaling $L_\textrm{ent}$ in methods UNC-\textbf{E}CD and UNC+\textbf{E}CD is empirically set to $\lambda = 1 / (360 \times |R^{l_\mathrm{abst}}|)$.

To provide quantitative results, we have evaluated Dice score coefficient and mean surface distance (MSD).
In an effort to efficiently utilize the available dataset, we have generated 5 hold-out experiments 
where the initial training set $D_\mathrm{an}$, the non-annotated set $D_\mathrm{pool}$, the validation set (all slices from 2 patients) and the test set (all slices from 9 patients) are randomly picked. 
All experiments are initially trained on 64 slices. 
For every active learning step, $n_\mathrm{rep}$\,$=$\,$32$ and $n_\mathrm{unc}$\,$=$\,$64$ is used. 
In Figs.~\ref{fig:scores} \&~\ref{fig:scores2}, we show the Dice score and MSD of different methods evaluated for the test set at 11 stages of active learning ranging from $4\%$ up to $27\%$ of the $D_\mathrm{pool}$. 
Conducted experiments are shown in two groups to increase clarity: (1) Comparison of our implementation of the baseline (UNC-ID) to uniform random sample querying (RAND) and sample querying based only on uncertainty (UNC) as seen in Fig.~\ref{fig:scores}; (2) Building on top of (1), improvements of ranking (UNC+ID) and the gain from $L_\mathrm{ent}$ during training and representativeness capabilities of $d_\mathrm{cont}$ for sample querying, UNC+\textbf{E}CD (c.f. Fig.~\ref{fig:scores2}).
In Fig.~\ref{fig:qualRes}, we show an example cross-section from a test volume, where segmentation superiority of our proposed method (UNC+\textbf{E}CD) when compared to baseline is already visible after a single active learning step.

We conducted one-sided paired-sample t-tests at the $5\%$ significance level on the mean Dice scores over all active learning steps for each hold-out experiment for UNC+\textbf{E}CD being superior to UNC-ID. Performance of UNC+\textbf{E}CD was statistically significantly better in 4 of 5 experiments.

\section{Discussions \& Conclusions}
\label{sec:discussions}

At early steps of active learning, one can see that the only uncertainty-based query sampling method (UNC) performs similar to random sample querying (RAND), with UNC only improving soon after $\approx$\,$12\%$ of $D_\mathrm{pool}$ is used in training (c.f. Fig.~\ref{fig:scores}).
While UNC-ID already yields better segmentation performance than just uncertainty-based sampling, by simply using ranking, one can see that the baseline method achieves a more substantial boost at early stages of active learning (see UNC+ID in Fig.~\ref{fig:scores2}).
This behavior suggests that the surrogate uncertainty metric can give a bad approximation when the trained data size is fairly low; i.e.,\ initial step(s).
However, the suboptimal segmentation performance gain can be compensated with representativeness, and even further improved when given a higher priority; i.e.,\ ranking instead of 2-step sample querying.

Upon combination of the proposed additional information maximization constraint during training and ranking combined with content distance at sample querying (UNC+\textbf{E}CD), we have observed the best Dice score on average at all active learning steps among the compared baseline and ranking extensions of the baseline methods.
Other possible combinations of our proposed extensions (UNC-CD, UNC+CD, UNC-\textbf{E}CD) yielded inferior performance to UNC+\textbf{E}CD, and hence are not included in the quantitative comparisons to reduce clutter. 

In this paper, we have comparatively studied the impact of different sample selection methods in active learning for segmentation. 
We have proposed 2 novel ways to query samples for active learning, which also can be combined to further boost performance during active learning steps. 
Compared to a state-of-the-art method, we have shown our proposed method to yield statistically significant improvement of segmentation Dice scores. 

\noindent\textbf{Acknowledgements.}
This work was funded by the Swiss National Science Foundation (SNSF), a Highly Specialized Medicine (HSM2) grant of the Canton of Zurich, and the EU’s 7th Framework Program (Agreement No. 611889, TRANS-FUSIMO).
We acknowledge NVIDIA GPU Grant support.

\bibliographystyle{splncs03}
\bibliography{bib}

\end{document}